\documentclass[runningheads]{llncs}

\newcommand\blfootnote[1]{%
  \begingroup
  \renewcommand\thefootnote{}\footnote{#1}%
  \addtocounter{footnote}{-1}%
  \endgroup
}

\usepackage{graphicx}

\begin{document}

\title{Improving Prostate Cancer Detection with Breast Histopathology Images}

\author{Umair~Akhtar~Hasan~Khan\inst{1} \and
Carolin~St\"urenberg\inst{2} \and 
Oguzhan~Gencoglu\inst{1} \and 
Kevin~Sandeman\inst{3} \and 
Timo~Heikkinen\inst{1} \and 
Antti~Rannikko\inst{4} \and 
Tuomas~Mirtti\inst{2} 
} 
\authorrunning{U. A. H. Khan, C. St\"urenberg, O. Gencoglu et al.}
%
\institute{Top Data Science Ltd., Helsinki, Finland\and 
University of Helsinki, Faculty of Medicine, Department of Pathology and Research Program in Systemic Oncology, Helsinki, Finland\and
Department of Laboratory Medicine, Department of Pathology, Sk{\aa}ne University Hospital, Malm{\"o}, Sweden \and
Helsinki University Hospital, Department of Urology and Research Program in Systemic Oncology, Faculty of Medicine, University of Helsinki, Helsinki, Finland
}
\maketitle 
\setcounter{footnote}{0}

\begin{abstract}
Deep neural networks have introduced significant advancements in the field of machine learning-based analysis of digital pathology images including prostate tissue images. With the help of transfer learning, classification and segmentation performance of neural network models have been further increased. However, due to the absence of large, extensively annotated, publicly available prostate histopathology datasets, several previous studies employ datasets from well-studied computer vision tasks such as ImageNet dataset. In this work, we propose a transfer learning scheme from breast histopathology images to improve prostate cancer detection performance. We validate our approach on annotated prostate whole slide images by using a publicly available breast histopathology dataset as pre-training. We show that the proposed cross-cancer approach outperforms transfer learning from ImageNet dataset{\blfootnote{This manuscript has been accepted to 15th European Congress on Digital Pathology (ECDP2019). \newline Corresponding Author : \texttt{oguzhan.gencoglu@topdatascience.com}}}.
\end{abstract}

\keywords{prostate cancer \and convolutional neural networks \and computer aided diagnosis \and breast cancer \and transfer learning.}

\section{Introduction}

Prostate cancer (PCa) is the second most common solid malignant disease among males in Western world and it derives from the glands within the prostate~\cite{siegel2017cancer}. The incidence of PCa is especially high in Northern America, Europe and most parts of Africa, and it is the second common cause of cancer-related deaths in western countries~\cite{bray2018global}. PCa is commonly found in older men over the age of 65 years, with a chance of 1 in 8 men diagnosed with the disease during their lifetime~\cite{siegel2017cancer}.

Histological examination of the surgical tissue and detection of cancer by a pathologist is still the gold standard in cancer diagnostics. PCa diagnostics is heavily time-consuming. Furthermore, it is based on subjective grading, i.e., there is considerable inter-pathologist variability in assessing the diagnosis. For instance, the study by Ozkan et al. reports that two pathologists disagreed on the presence of cancer in 31 out of 407 biopsy cores and the overall concordance of the assessed Gleason scores was only 51.7\%, depicting the challenges in diagnosing PCa consistently~\cite{ozkan2016interobserver}. Therefore, development of computer-aided decision support tools is crucial for saving time, increasing precision and enhancing standardisation in diagnostics for pathologists.

There has been substantial interest in developing digital image processing and machine learning-based methods for automatic analysis of pathology images in order to perform tissue classification and disease grading, as well as predicting disease outcome and enhancing precision medicine~\cite{madabhushi2016image}. Specifically, recent advancements in machine learning research involving deep neural networks, i.e., deep learning, have successfully increased the performance of such analyses~\cite{hu2018deep}. However, proposed deep learning models often require significant amount of annotated data in order to be successfully trained. As cohort sizes can be small and the annotation of histopathology images is very time consuming, a concept called \textit{transfer learning}, i.e., training a neural network with an external dataset and then fine-tuning the model with the dataset at hand, may prove beneficial. Such an approach of fine-tuning a pre-trained model has been shown to outperform training the same neural network architecture from scratch in studies involving analysis of digital pathology images~\cite{kieffer2017convolutional,mehra2018breast,mormont2018comparison}. Transfer learning may also be beneficial for adapting to domains in which images are obtained with different microscopes or staining procedures.

In this work, we propose a cross-domain transfer learning approach, specifically from breast histopathology images to prostate histopathology images, in order to train a deep convolutional neural network (CNN) for the detection of cancerous regions in PCa whole slide images (WSIs). From the pathological point of view, breast cancer (BrCa) and PCa are both adenocarcinomas (glandular origin) and the most common cancers among the respective genders. The rationale for this approach is that the cellular composition of BrCa and PCa have more visual similarity than the images in conventional pre-training materials, such as ImageNet dataset~\cite{deng2009imagenet}, applied in earlier studies. Based on this hypothesis, we propose a cross-domain transfer learning scheme between the images of two types of cancers. We show that pre-training a neural network model on BrCa histopathology images and fine-tuning it with PCa histopathology images increases the performance compared to training the model from scratch. In addition, we show that this approach outperforms models pre-trained on ImageNet dataset which has been the standard dataset for transfer learning models in deep learning-based digital pathology analysis. The main focus of this work has not been to maximize detection performance through rigorous data augmentation, neural architecture search, hard negative mining, hyper-parameter optimization or model ensembling but rather to propose a cross-cancer transfer learning alternative to ImageNet dataset. To the best of our knowledge, this study is the first study to propose a cross-domain (breast tissue to prostate tissue) transfer learning scheme for deep learning based PCa diagnosis.

\section{Related Work}

There have been several studies utilising transfer learning, especially with CNNs, to detect, classify, segment cancerous regions or to predict the Gleason grade in PCa histopathology images~\cite{arvaniti2018coupling,arvaniti2018automated,campanella2018terabyte,isaksson2017semantic,kallen2016towards,nagpal2018development,schaumberg2018h}. A typical approach recurring in previous studies is to divide the image into smaller tiles/patches (overlapping or non-overlapping) and to perform binary or multi-class classification of the tiles. Reconstruction of tile-level or pixel-level probability map of a given class for the original image is similarly performed in a sliding window fashion using the inference results of the tiles. Tile dimensions (in pixels) as well as the dimensions that are fed into a CNN vary between studies, e.g., 250x250~\cite{isaksson2017semantic}, 400x400 downscaled to 224x224~\cite{arvaniti2018coupling}, 512x512 downscaled to 256x256 and further cropped to 224x224~\cite{schaumberg2018h}, 750x750 downscaled to 250x250~\cite{arvaniti2018automated}, 911x911~\cite{nagpal2018development}.

One common transfer learning approach is to use an architecture that has performed well in other tasks (e.g. object detection in natural images) and to train it from scratch. Such an approach has been utilized by different works~\cite{isaksson2017semantic,nagpal2018development}. For Gleason grading, Nagpal et al.~\cite{nagpal2018development} employed an architecture that has been shown to reach significant performance on well-known ImageNet dataset~\cite{deng2009imagenet}, i.e., InceptionV3~\cite{szegedy2016rethinking} and the study by Isaksson et al.~\cite{isaksson2017semantic} proposes a U-net~\cite{ronneberger2015u} based semantic segmentation of prostate tissue.

Another transfer learning method is to use a pre-trained model as a feature extractor and perform further classification with a separate classifier. This is achieved by extracting the representations out of the intermediate layers of a pre-trained network. This approach has been used to predict Gleason score by extracting features from different layers of the 22-layer OverFeat architecture~\cite{sermanet2013overfeat} (pre-trained on ImageNet) and feeding the features into random forest and support vector machine classifiers~\cite{kallen2016towards}.

Finally, the most prevalent way to perform transfer learning is to employ a pre-trained model and to fine-tune it with the data at hand. Several fine-tuning approaches can be utilised such as fine-tuning all the layers, freezing the initial neural network layers (usually the convolutional layers) and fine-tuning only the last few layers or sequential layer-wise fine-tuning~\cite{arvaniti2018coupling,arvaniti2018automated,campanella2018terabyte,schaumberg2018h}. Used architectures for this purpose include either original implementations or implementations with small modifications of the following: AlexNet~\cite{krizhevsky2012imagenet} in~\cite{campanella2018terabyte}, VGG~\cite{simonyan2014very} in~\cite{arvaniti2018automated,campanella2018terabyte}, ResNet~\cite{he2015deep} in~\cite{arvaniti2018coupling,arvaniti2018automated,campanella2018terabyte,schaumberg2018h}, InceptionV3~\cite{szegedy2016rethinking} in~\cite{arvaniti2018automated}, MobileNet~\cite{howard2017mobilenets} in~\cite{arvaniti2018automated} and DenseNet~\cite{huang2017densely} in~\cite{arvaniti2018automated}. 

Even though the domains are considered both visually and in nature very different (natural images vs. prostate tissue images), most of the transfer learning schemes use architectures or models trained on ImageNet dataset. This is due to the absence of publicly available, large-scale, extensively annotated PCa histopathology datasets. In addition, the high number of images (over 1.2 million images with 1000 classes) and availability of several CNN models pre-trained on it, renders ImageNet a prominent dataset for the basis of transfer learning.

The performance of the deep neural network models in the abovementioned studies varies depending on the overall task at hand, dataset used, evaluation setup (sampling, cross-validation, training/validation/test splitting etc.), whether data augmentation was used or not and whether an ensemble of several classifiers was used or not. Therefore, fair comparison between studies is a non-trivial task. Most frequently used performance metric for reporting tile-level classification is \textit{area under the receiver operating characteristic curve} (AUC)~\cite{campanella2018terabyte}.

\section{Methods}
\subsection{Data}

\begin{figure}
\centering
\includegraphics[width=0.68\columnwidth,trim={0 0.4cm 0 1.5cm},clip]{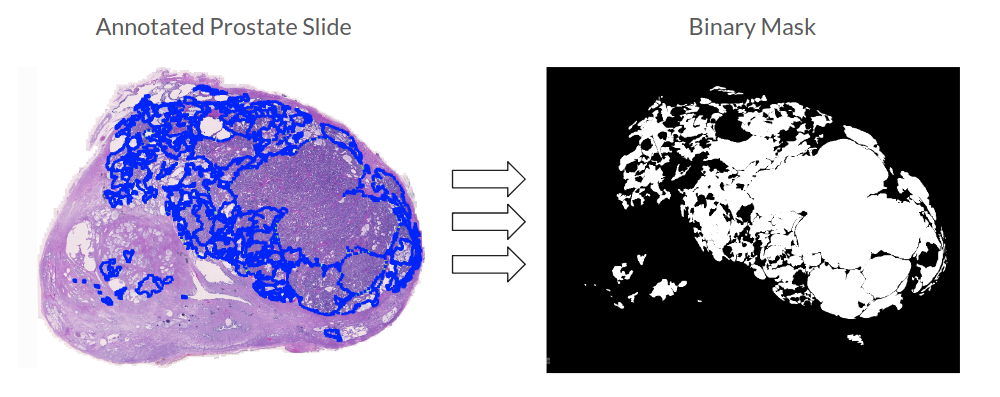}
\caption{An example of WSI with its annotations and corresponding binary mask.} 
\label{fig1}
\end{figure}

Here, we aim to utilize a well known image dataset ImageNet and a previously annotated BrCa dataset, Cancer Metastases in Lymph Nodes Challenge 2016 (CAMELYON16)\footnote{\url{https://camelyon16.grand-challenge.org/}}, to improve cancer detection with CNNs in our PCa database. The dataset of 28 macro (2 inch x 3 inch) histological surgical specimen WSIs was prepared from 28 patients with clinically relevant PCa (Gleason score $\geq$ 6) who had undergone prostatectomy during the years 2014 or 2015 in the Helsinki University Hospital, Helsinki, Finland. The slides were stained with H\&E staining in a clinical-grade laboratory (HUSLAB Laboratory Services) at the Helsinki University Hospital. The scanning of the WSIs was performed by Zeiss Axio Scan.Z1 at a resolution of 0.220 $\mu$m x 0.220 $\mu$m per pixel. Cancerous loci were annotated with polygons using the open source Automated Slide Analysis Platform\footnote{\url{https://github.com/computationalpathologygroup/ASAP}} (ASAP) software at 750 $\mu$m magnification. Annotation of a single slide took 0.5 to 6 hours depending on the slide, resulting in an average of 3 hours per slide. An example WSI and corresponding annotations are shown in Figure~\ref{fig1}. Minimum, mean and maximum cancerous area percentages with respect to the image size are 0.7\%, 7.4\% and 29.1\%, respectively. Minimum and maximum number of polygon annotations (corresponding to the cancerous area/region in an image) are 4 and 208, respectively.

For pre-training, publicly available CAMELYON16 dataset was employed with 110 WSIs with nodal metastases verified by H\&E staining~\cite{bejnordi2017diagnostic}. In this dataset, WSIs have been acquired by 2 different scanners, i.e., Pannoramic 250 Flash II - 3DHISTECH and NanoZoomer-XR Digital slide scanner C12000-01 - Hamamatsu Photonics with resolutions of 0.243 $\mu$m x 0.243 $\mu$m and 0.226 $\mu$m x 0.226 $\mu$m per pixel, respectively~\cite{bejnordi2017diagnostic}.

\subsection{Classification and Transfer Learning}

We divided 28 WSIs of PCa, each corresponding to a single patient, into training and held-out test sets with 22 and 6 images, respectively. Each image is then divided into non-overlapping tiles of 256x256 pixels to be fed into CNNs. From the training set, we randomly sampled 300,000 cancerous tiles and 300,000 non-cancerous tiles (white background is not sampled) in order to ensure a 50\%-50\% class balance for binary classification (in total 600,000 tiles). Randomness in the explained procedure is fixed for every experiment in order to ensure the exact same sampling and data splits.

\begin{figure}
\centering
\includegraphics[width=0.9\columnwidth,trim={3.8cm 1.6cm 0 0.5cm},clip]{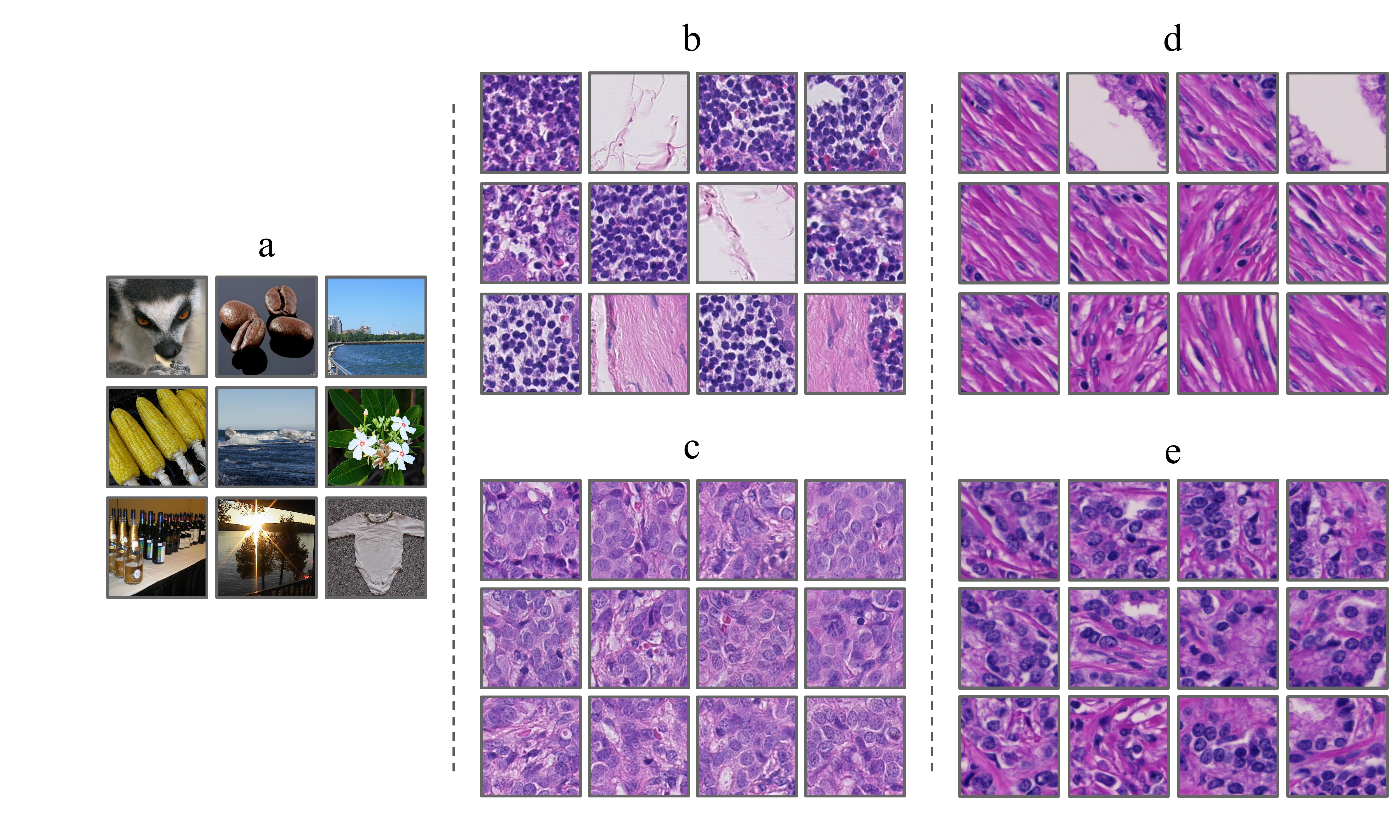}
\caption{Examples of data from a. ImageNet b. benign breast tissue c. cancerous breast tissue d. benign prostate tissue e. cancerous prostate tissue.} \label{fig2}
\end{figure}

For classification, we used an InceptionV3 architecture~\cite{szegedy2016rethinking}, i.e., the convolutional backbone of the well-known architecture followed by 2 fully-connected layers with 512 and 128 units, respectively and a single unit output layer. Dense layers employed \textit{ReLU} activation functions and a dropout rate of 0.8. Output layer employed a \textit{sigmoid} activation. Loss function is chosen to be \textit{binary crossentropy} and the optimizer is chosen to be \textit{Adam} with a learning rate of $10^{-4}$. Three different models were trained with the same architecture and same training data: training from scratch (random weight initialization), fine-tuning on ImageNet pre-trained model, fine-tuning on BrCa pre-trained model. Only the convolutional layer weights were used from pre-trained models (fully-connected layers are still randomly initialized). Example data used during the training of the models can be seen in Figure~\ref{fig2}. For BrCa pre-training, 110 WSIs (no held-out test set) from CAMELYON16 dataset were used with a total of 500,000 randomly sampled tiles (again 50\%-50\% target distribution). For all PCa and BrCa models, a random split of 80\%-20\% was employed for training and validation data, respectively. Each training was run for 50 epochs and the model weights reaching lowest validation error in that particular training were saved. Finally, models were evaluated on the 6 held-out test PCa WSIs and tile-level as well as pixel-level AUC scores were calculated. 

\section{Results and Discussion}

Results of the experiments can be examined from Table~\ref{tabel1}. We show that pre-training the model on breast tissue samples and fine-tuning it with prostate tissue samples improves the tile-level classification AUC score by 0.051 from 0.873 to 0.924. Similarly, pixel-level AUC score increases from 0.879 to 0.936. In addition, we compare the performance of pre-training on BrCa data with pre-training on ImageNet data. We show that the pre-training on BrCa images outperform pre-training of ImageNet with a 2.3\% improvement (0.903 to 0.924) in tile-level and 2.2\% improvement (0.916 to 0.936) in pixel-level AUC score.

\begin{table}
\caption{Tile-level and pixel-level AUC scores of the trained CNNs with different pre-training data, evaluated on the 6 test slides each belonging to an individual prostate cancer patient.}\label{tabel1}
\centering{
\begin{tabular}{|c|c|c|c|c|c|c|c|c|}
\cline{2-9}
\multicolumn{1}{c|}{} & Pre-training & Slide 1 & Slide 2 & Slide 3 & Slide 4 & Slide 5 & Slide 6 & Overall \\
\hline
Tile-level & None & 0.859 & 0.964 & 0.902 & 0.831 & 0.794 & 0.849 & 0.873 \\
Tile-level & ImageNet & 0.898 & 0.952 & 0.933 & 0.932 & 0.854 & 0.881 & 0.903\\
Tile-level & CAMELYON16 & 0.916 & 0.971 & 0.946 & 0.953 & 0.885 & 0.874 & \textbf{0.924}\\
\hline
Pixel-level & None & 0.861 & 0.973 & 0.911 & 0.835 & 0.792 & 0.882 & 0.879\\
Pixel-level & ImageNet & 0.904 & 0.970 & 0.942 & 0.938 & 0.859 & 0.915 & 0.916\\
Pixel-level & CAMELYON16 & 0.920 & 0.979 & 0.955 & 0.958 & 0.890 & 0.912 & \textbf{0.936}\\
\hline
\end{tabular}
}
\end{table}

Contributions of this work lie in the transfer learning paradigm which has been shown to be beneficial to the model performance in several studies involving digital pathology analysis with deep neural networks~\cite{kieffer2017convolutional,mehra2018breast,mormont2018comparison}. Due to the absence of a large, publicly available, extensively annotated prostate histology image dataset, transfer learning inside the same domain has not been possible so far. This led to frequent use of ImageNet dataset for this purpose~\cite{arvaniti2018coupling,arvaniti2018automated,campanella2018terabyte,schaumberg2018h}. Our results bolster the intuition behind this practice, i.e., first-layer representations learned by deep neural networks are not specific to a particular dataset but applicable to many even though the tasks are visually different. However, our study proposes an alternative to ImageNet pre-training by utilising a large dataset of breast WSIs. Our experiment results show evidence of enhanced knowledge transfer due to visual similarities of the two cancer domains which is lacking in natural images of objects, i.e., ImageNet. In addition, such cross-domain transfer learning may also improve the generalization ability of the models to different scanners, image resolutions and stainings. 

As our methodology can be generalized to other cancer domains, future work includes extensive analysis of cross-domain pre-training from different cancer pathology images, with varying neural network architectures and training schemes. In addition, double pre-training scheme will be examined in which a model can be first trained on ImageNet (or on a large dataset of similar nature), followed by a fine-tuning with breast histopathology images and then finally further fine-tuned with the data at hand. 

\section{Conclusion}

In this work we propose a cross-domain, deep convolutional neural network-based transfer learning scheme, specifically from breast to prostate histopathology images, to enhance prostate cancer detection performance. In addition, we compare the proposed breast histopathology pre-training with the well-known ImageNet dataset pre-training. Our results show that the model pre-trained on breast cancer images, further fine-tuned with prostate cancer images performs better than the model that is trained from scratch or pre-trained on ImageNet dataset. We believe our study serves as an advancement in the field of machine learning-based analysis of prostate cancer histopathology images by providing evidence for a transfer learning scheme between different cancer domains.

\bibliographystyle{splncs04}
\bibliography{references}

\end{document}